\documentclass{article}
\usepackage{spconf,amsmath,graphicx}
\usepackage{amsfonts,dsfont}
\usepackage{bm}
\usepackage{amssymb}
\usepackage{enumerate}
\usepackage{balance}
\usepackage{geometry}
\title{IMAGE RETRIEVAL WITH HIERARCHICAL MATCHING PURSUIT}

\name{Shasha Bu, \qquad Yu-Jin Zhang
\thanks{This work was supported by National Nature Science Foundation
(NNSF: 61171118) and Specialized Research Fund for the Doctoral
Program of Higher Education (SRFDP-20110002110057).}
}
\address{Department of Electronic Engineering,
\\Tsinghua University,
\\Beijing 100084, China
\\Email: {boss12@mails.tsinghua.edu.cn, \quad zhang-yj@mail.tsinghua.edu.cn}
}

\begin{document}
%
\maketitle
\begin{abstract}
A novel representation of images for image retrieval is introduced in this paper, by using a new type of feature with remarkable discriminative power. Despite the multi-scale nature of objects, most existing models perform feature extraction on a fixed scale, which will inevitably degrade the performance of the whole system. Motivated by this, we introduce a hierarchical sparse coding architecture for image retrieval to explore multi-scale cues. Sparse codes extracted on lower layers are transmitted to higher layers recursively. With this mechanism, cues from different scales are fused. Experiments on the Holidays dataset show that the proposed method achieves an excellent retrieval performance with a small code length.

\end{abstract}


\section{Introduction}
\label{sec:intro}

Image retrieval has been increasingly popular in recent years. Searching images such as pictures of a scenic spot or an animal has become a part of everyday life for many people, either from the internet or database in hand. However, with image database growing increasingly larger, how to find the intended images from so many images is a problem presented in image retrieval. A lot of works have been done in this field \cite{zheng2013visual}\cite{zheng2014bayes}\cite{graph}\cite{zheng2014coupled}\cite{zheng2014seeing}.

Recent works on image retrieval mainly concentrate on content based image retrieval (CBIR). Features from images are extracted and compared for similarity measurement based on which the most similar images to the query are returned.

Bag-of-features (BoF) model \cite{sivic2003video} is extensively used in CBIR which often obtains good performance. Methods following such a framework often use Scale-invariant feature transform (SIFT) \cite{lowe1999object}, which is robust against many image transformations. However, the vector quantization (VQ) \cite{gray1984vector} in BoF model only assumes that each feature is related to a single visual word, and thus ignores the correlation between the feature and other words. What is more, SIFT is a local feature which is unable to capture the global cues. And features of the same image are irrelevant to each other, limiting the fusion of cues between them. Sparse coding techniques and global features have been proposed to fix the problem \cite{shi2012sift}\cite{jegou2010aggregating}\cite{thiagarajan2012supervised}\cite{perronnin2010large}\cite{zhengtopic}\cite{liu2012discriminant}\cite{liu2013learning}. Nevertheless, neither utilizing one-layer sparse coding nor leveraging global feature on a fixed scope can cues of different scales be adequately explored. The success of hierarchical matching pursuit (HMP) algorithm in classification \cite{bo2013multipath} motivates us to employ the hierarchical sparse coding architecture in image retrieval to explore multi-scale cues.

A global feature using HMP is introduced in this paper for image retrieval, which has not been considered in this field to our knowledge. The global cues as well as features on different scales are extracted, forming a sparse representation. Images are first partitioned into patches of different sizes. Then, sparse codes are extracted from smaller patches and spatially pooled on larger patches recursively. Finally, a hierarchical sparse coding architecture is constructed, and sparse representations extracted from the hierarchical layers are adopted for retrieval. Experiments conducted on the Holidays dataset \cite{jegou2008hamming} demonstrate the effectiveness of the proposed approach, where excellent performance compared with prior methods is obtained.

\section{Sparse Coding in CBIR}
\label{sec:sc}

This section presents the procedure of utilizing sparse coding for CBIR. A standard sparse coding model can be formulated as follows. Given an over completed codebook $\mathbf{C}$ ($ \mathbf{C} \in \mathds{R}^{D \times K}$) and a basic feature $\mathbf{y}$ ($\mathbf{y} \in \mathds{R}^{D}$), a vector $\mathbf{x}$ ($\mathbf{x} \in \mathds{R}^K$) with sparsity $L$ is generated to approximate $\mathbf{y}$ \cite{thiagarajan2012supervised} as
\begin{equation}\label{equ:sc}
\begin{aligned}
&\min\limits_{\mathbf{x}} {{\|\mathbf{y} - \mathbf{Cx}\|}^2}, s.t. \|\mathbf{x}\|_0 \leq L.
\end{aligned}
\end{equation}
Orthogonal matching pursuit (OMP) \cite{bo2013multipath} is usually employed to solve Eq. \eqref{equ:sc}.

When sparse coding is used in CBIR, features are extracted from the image and sparsely coded using Eq. \eqref{equ:sc}. Then, max-pooling \cite{bo2013multipath} is applied to all sparse codes of the image to form a sparse representation which is used for similarity measurement in the search step.

The BoF model can also be treated as a special case of sparse representation \cite{wang2010locality}. Low-level features extracted from the image are quantized to the nearest visual words in the codebook using VQ as
\begin{equation}\label{equ:bof}
\begin{aligned}
\min\limits_{\mathbf{x}} {{\|\mathbf{y} - \mathbf{Cx}\|}^2}, s.t. \|\mathbf{x}\|_0 = 1, \|\mathbf{x}\|_1 = 1, \mathbf{x}(i) \geq 0, \forall i.
\end{aligned}
\end{equation}

Codes of all features of an image are aggregated using average pooling \cite{shi2012sift}, generating a final sparse representation of the BoF model.
Note that Eq. \eqref{equ:bof} only allows a sparsity level 1 of vector $\mathbf{x}$ which means a feature is assigned to only one visual word in the codebook in a hard manner. However, this may not be appropriate since a feature could also be related to multiple visual words, which has been proved in \cite{shi2012sift}, and thus the retrieval performance of BoF is limited while OMP can be utilized to improve it by assigning a feature to more visual words.

\section{proposed approach}
\label{sec:proposedapproach}

This section describes the hierarchical matching pursuit for image retrieval approach (HMP-IR). The correlations with multiple visual words are explored using OMP, and discriminative features of different scales are extracted using hierarchical sparse coding layers. Global cues can also be utilized by max pooling on spatial pyramids. A three-layer architecture of the whole HMP-IR algorithm is shown in Fig. \ref{fig:HMP-IR}. We use the same parameter settings as \cite{bo2013multipath}. More details are shown below.

\begin{figure}[!t]
\centering
\includegraphics[width=8cm]{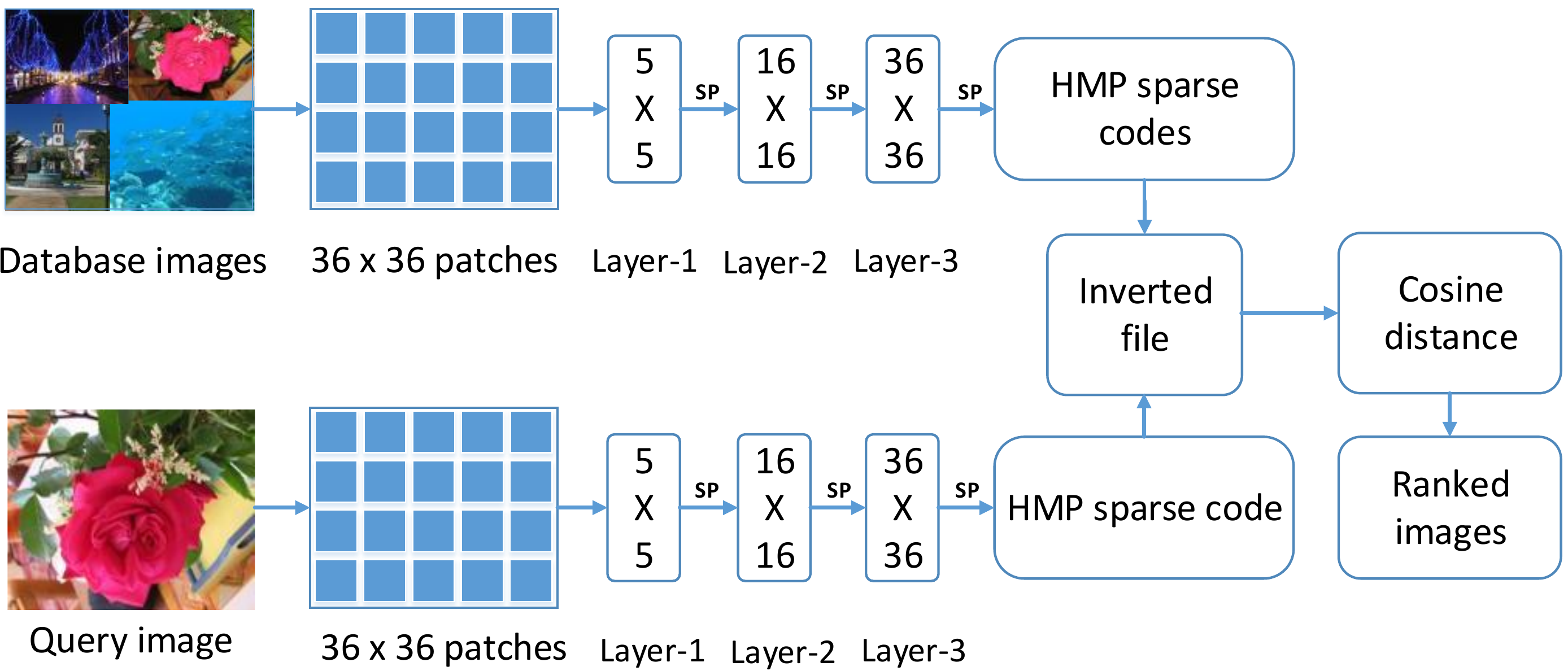}
\caption{Architecture of a three-layer hierarchical matching pursuit. Spatial max pooling is denoted by SP.}
\label{fig:HMP-IR}
\end{figure}

\subsection{Extracting HMP Representation}
\label{ssec:featract}

This subsection shows how to form a sparse HMP representation for a given image. The HMP representation consists of multiple layers. Input data of the first layer are raw patches sampled from images, and input of the higher are the pooled sparse codes from the previous layer. Sparse codes are extracted and pooled recursively on different layers. Mutual incoherence KSVD (MI-KSVD) method is adopted for codebook training \cite{bo2013multipath}. A spatial pyramid is constructed on the final layer. The coding procedure for a three-layer HMP-IR is as follows.

{\bf The first layer:} Sparse codes from small patches are extracted and adopted for generating representations for mid-level patches. A mid-level patch $P$ (\textit{e.g.} 16x16) is further divided into small spatial cells, and each cell is divided into small image patches (\textit{e.g.} 5x5) with overlaps. A sparse code is extracted from each small patch using the codebook of this layer. Codes of small patches within a cell $Ce$ are aggregated using max-pooling as
\begin{equation}\label{equ:maxpooling}
\begin{aligned}
F(Ce) = \max_{j\in Ce}[&\max(x_{j1},0),...,\max(x_{jM},0),\\
&\max(-x_{j1,0}),...,\max(-x_{jM,0})],
\end{aligned}
\end{equation}
where $j$ is the index of a small patch within the cell $Ce$, and $x_{jm}$ is the $m$-th element of the j-th sparse code vector $x_{j}$ in cell $Ce$. The positive and negative elements of vector $x_{j}$ are split into separate features and weighted differently by the higher layer encoder. Feature $F_{P}$ of mid-level patch $P$ is the concatenation of codes of all spatial cells $Ce_{s}^{P}, s = [1,2,...S]$ in $P$ as
\begin{equation}\label{equ:featureP}
F_{P} = [F(Ce_{1}^{P}),..,F(Ce_{2}^{P}),...,F(Ce_{S}^{P})].
\end{equation}
The feature $F_{P}$ is then $\ell_2$-normalized \cite{bo2013multipath} and fed to the second layer.

{\bf The second layer:} The features $F_{P}$ from the first layer are delivered to the second layer and processed the same way as raw patches on the first layer. Sparse codes for each feature $F_{P}$ are drawn and spatially max-pooled within each cell. Codes of each cell are concatenated on large image patches (\textit{e.g.}, 36x36). Then, the concatenated features on large image patches are normalized and transmitted to the third layer.

\begin{figure}[!t]
\centering
\includegraphics[width = 8cm]{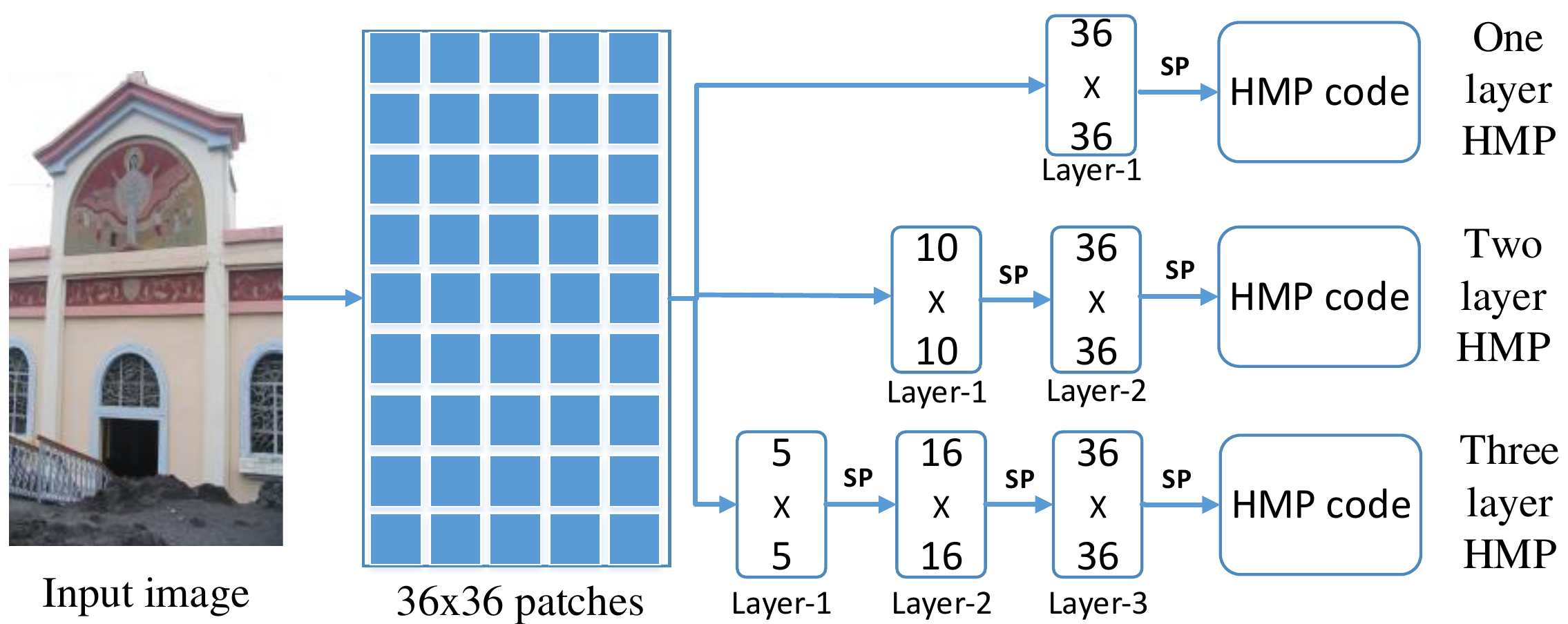}
\caption{Procedures of three different HMP-IR methods. SP indicates spatial max pooling.}
\label{fig:three_basic_layers}
\end{figure}

{\bf The third layer:} The features generated from the second layer are sparsely coded on the third layer. On this final layer, max pooling on spatial pyramids on the whole image is conducted. The pooled descriptors are $\ell_2$ normalized to form a sparse representation for the whole image. The coding procedures for the three different HMP-IR methods are illustrated in Fig. \ref{fig:three_basic_layers}.

\subsection{HMP Representation for Image Retrieval}
\label{ssec:HMPforRetrieval}

Representations of the database images computed in Sec. \ref{ssec:featract} are sparse, and are utilized for generating an inverted file \cite{sivic2003video} to speed up the searching procedure. In the search step, the query is coded in the same way, and then the inverted file is used to identify the candidate images. Cosine distance \cite{sivic2003video} is employed to evaluate the similarities between the candidates and the query.

\section{experiments}
\label{sec:experiments}

In this section, performances of the proposed approach utilizing different numbers of layers are presented. Comparisons with the BoF model utilizing RootSIFT features \cite{arandjelovic2012three} and other image retrieval methods are conducted on different code lengths. RootSIFT features are produced from SIFT features and perform better than the latter. The mean Average Precision (mAP) is adopted to evaluate different methods.

\subsection{Parameter Settings}
\label{ssec:parameters}

Two groups of HMP-IR methods are utilized to evaluate the performance on the Holidays dataset \cite{jegou2008hamming} with three different numbers of layers. In each group, one-layer HMP-IR (HMP-IR1), two-layer HMP-IR (HMP-IR2) and three-layer HMP-IR (HMP-IR3) methods are implemented on 36x36 image patches. Codebook sizes of each group on the final layer are set to 500 and 1000, respectively, to test the influence of codebook size on retrieval performance.

On the final layer, image-level features are obtained by max pooling on spatial pyramids on the whole image. Parameters of spatial pyramids are set to 1x1, 2x2 and 3x3 on the whole image. Different combinations of them are implemented. Note that the length of descriptor before spatial max pooling is double the size of the codebook on the final layer because of pooling in Eq. \eqref{equ:maxpooling}.

We adopt the BoF model \cite{zheng2013lp} as baseline. An $\ell_p$-norm inverse document frequency (IDF) \cite{zheng2013lp} weighting strategy ($p=3$) is employed to obtain a higher result.

\subsection{Retrieval Results on the Holidays Dataset}
\label{ssec:simulation}

\begin{figure}[!t]
\centering
\includegraphics[width=8cm]{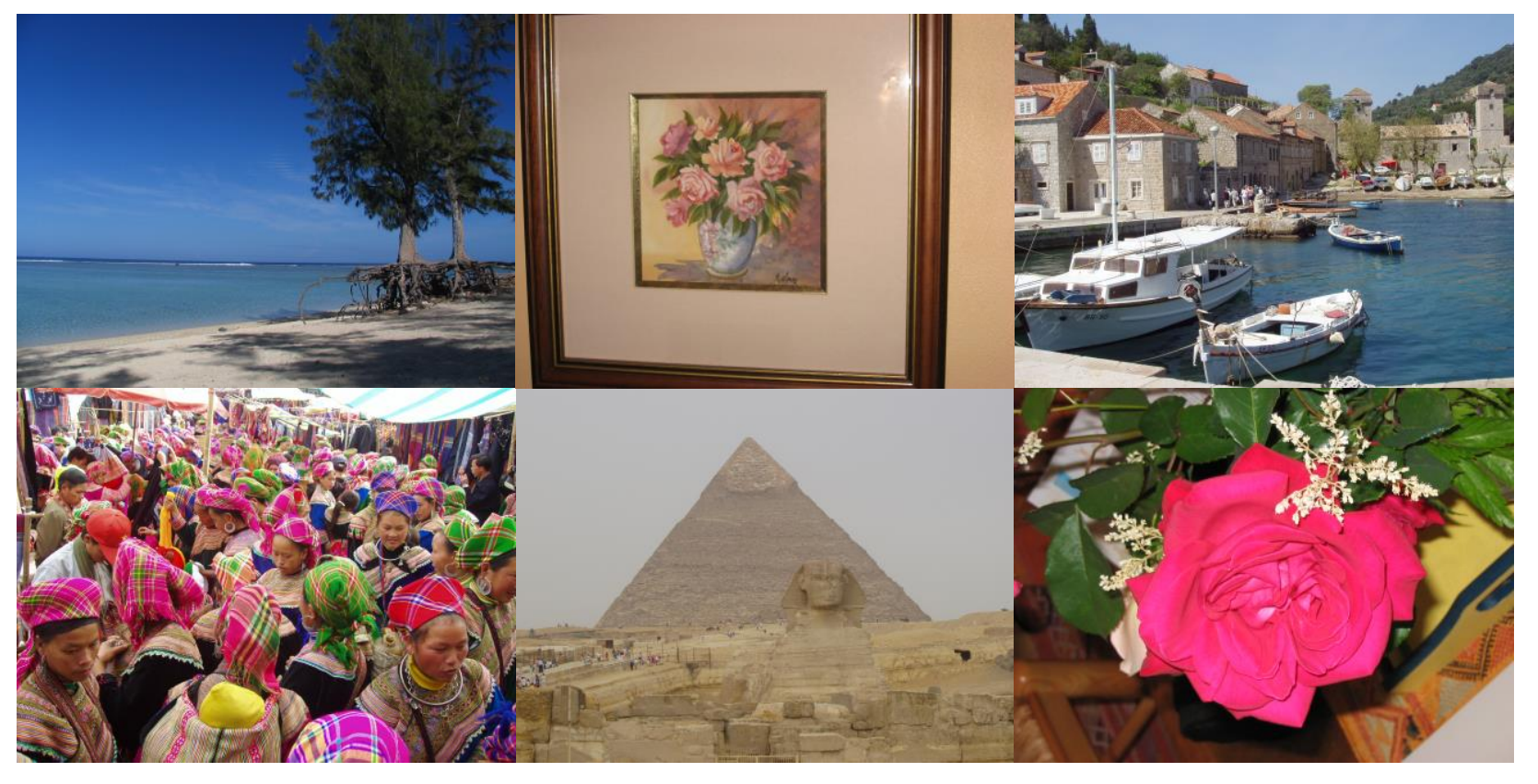}
\caption{A few examples on the Holiday dataset.}
\label{fig:holiday_samples}
\end{figure}

The Holidays dataset is widely used in image retrieval and contains 1491 color images taken on a large variety of scenes with 500 queries \cite{jegou2008hamming}. A few example images are shown in Fig. \ref{fig:holiday_samples}.

Comparison of the proposed HMP-IR2 method (pooled on 1x1 pyramid) with BoF and other state of art methods such as vector of locally aggregated descriptors (VLAD) \cite{jegou2010aggregating} and Fisher \cite{perronnin2010large} is presented in Table \ref{tab:diffdimension}. Codebook size is denoted by $K$. The final length of the feature is denoted by $D$. Results from Table \ref{tab:diffdimension} show that the HMP-IR method outperforms the others with a shorter code. The storage is reduced from 365MB to 6.63MB compared with BoF. Query time for each method are 0.0587s and 0.0554s, respectively. The query time doesn't decrease because a single feature is assigned to more visual words in HMP-IR, and thus more candidates are selected for similarity measurement.

\begin{table}[!t]
\caption{Comparison of different methods on the Holidays dataset.}
\label{tab:diffdimension}
\centering
\begin{tabular}{c|c|c|c}
\hline
  Methods & $K$ & $D$ & mAP \\ \hline\hline
  BoF\cite{zheng2013lp} & 20 000 & 20 000 & 0.4713 \\ \hline
  VLAD\cite{jegou2010aggregating} & 64 & 8192 & 0.526 \\ \hline
  Fisher\cite{perronnin2010large} & 64 & 4096 & $\mathbf{0.595}$ \\ \hline\hline
  HMP-IR2 & 1000 & 2000 & $\mathbf{0.6822}$ \\ \hline
\end{tabular}
\end{table}

\begin{figure}[!t]
\centering
\includegraphics[width=8cm]{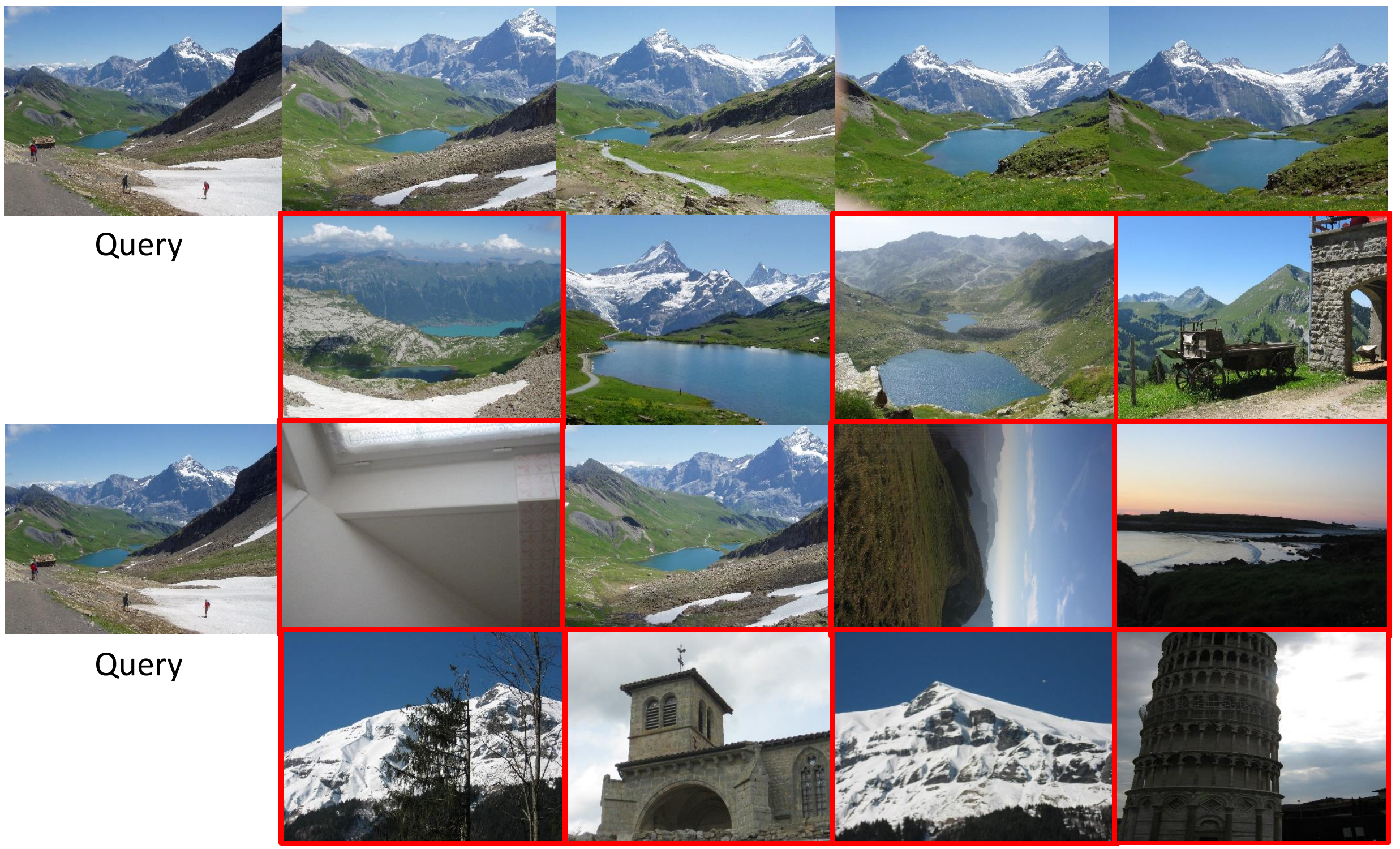}
\caption{The top 8 images returned by HMP-IR2 and BoF. The first and second rows correspond to HMP-IR2, and the lower rows to BoF. Incorrect results are marked with red boxes. The mAP for each are 0.8012 and 0.0616, respectively. Number of ground truth is 8.}
\label{fig:query_results}
\end{figure}

As is shown in Fig. \ref{fig:query_results}, the HMP-IR2 extracts discriminative features from multiple scales (the small-scale blue river and the grass land and mountain of large scale), while BoF mainly learns features of fixed scale (the large-scale white road and mountain) which take more area of the image than others.

Performances of the two groups of HMP-IR methods are shown in Table\ref{tab:pyramid} with max pooling on 1x1 pyramid. The final codebook sizes ($K$) are 500 and 1000, respectively.

\begin{table}[!t]
\caption{Performances of three HMP-IR methods with different codebook sizes ($K$) on 1x1 pyramid.}
\label{tab:pyramid}
\newcommand{\Rown}{\stepcounter{Rownumber}\theRownumber}
\centering
\begin{tabular}{c|c|c|c} \hline
   $mAP$ & HMP-IR1 & HMP-IR2 & HMP-IR3 \\ \hline\hline
   $K$ = 500 & 0.4849 & $\mathbf{0.6537}$ & 0.6390 \\ \hline
   $K$ = 1000& 0.4992 & $\mathbf{0.6882}$ & 0.6603 \\ \hline
\end{tabular}
\end{table}

\begin{figure}[!t]
\centering
\includegraphics[width = 8cm]{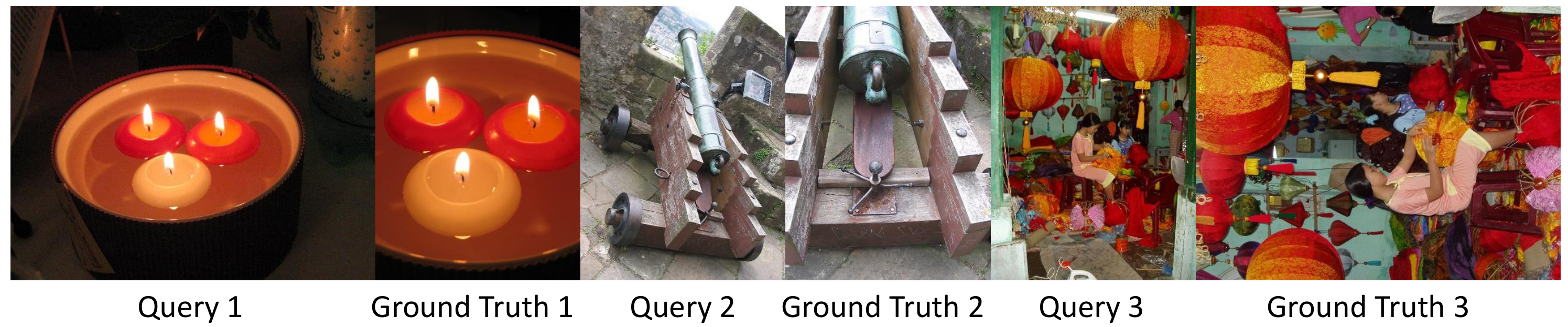}
\caption{Three failure examples of HMP-IR3 ($K=1000$), on 1x1 pyramid. Queries and the corresponding ground truths are shown in each group.}
\label{fig:fails-H3}
\end{figure}

According to  Table \ref{tab:pyramid}, performance is improved with a larger codebook since more cues can be encoded. HMP-IR2 and HMP-IR3 outperform HMP-IR1 which proves that the correlations between visual words are excavated by delivering codes between different hierarchical layers, and cues of image are thoroughly used, which is shown in Fig. \ref{fig:query_results}. Performance of three-layer HMP-IR is not as good as two-layer HMP-IR. A few failure cases of HMP-IR3 with 1000 codebook size are shown in Fig. \ref{fig:fails-H3}, where the ground truths are of different view points from the queries. This may indicate that angle cues are lost through too many layers of sparse coding.

\begin{figure}[!t]
\centering
\includegraphics[width=8cm]{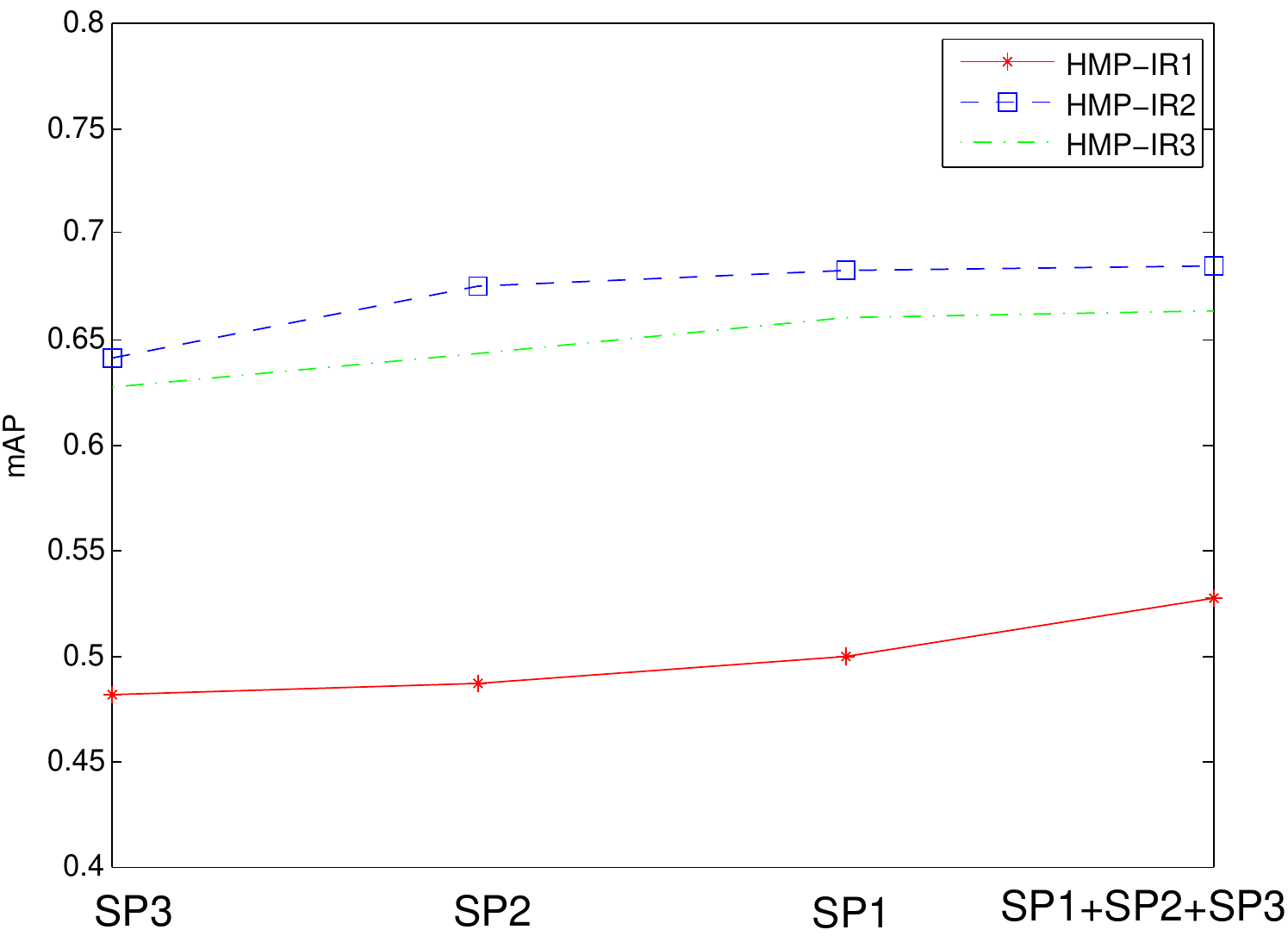}
\caption{Performance of three HMP-IR methods ($K=1000$) on different pyramids. SP1, SP2, SP3 indicates a pyramid scale of 1x1, 2x2 and 3x3, respectively. SP1+SP2+SP3 denotes the combination of three pyramids.}
\label{fig:HMP-IR1K}
\end{figure}

Fig. \ref{fig:HMP-IR1K} shows the performance of the second group of HMP-IR methods on different pyramids. SP1, SP2 and SP3 denotes 1x1, 2x2 and 3x3 spatial pyramid on the whole image, respectively. SP1+SP2+SP3 indicates the combination of three pyramids. It can be drawn from Fig. \ref{fig:HMP-IR1K} that better performance is obtained on larger grid (\textit{e.g.}, 1x1) which is easy to understand as pooling on larger grid can embed more spatial cues.
\balance

\section{Conclusion}
\label{sec:conclusion}
In this paper, we introduce the hierarchical matching pursuit method from image classification and modify the procedure to apply it to image retrieval. Multi-scale features are fused, and global cues are explored to obtain a better performance. Experiments show that our approach outperforms many other methods with a shorter descriptor. Future works include testing the scalability on large scale and different datasets and fusion with other features.

%
{
\small
\bibliographystyle{IEEEbib}
\bibliography{refs}

\begin{thebibliography}{10}

\bibitem{zheng2013visual}
L.~Zheng and S.~Wang,
\newblock ``Visual phraselet: Refining spatial constraints for large scale
  image search,''
\newblock {\em IEEE Signal Process. Lett.}, vol. 20, no. 4, pp. 391--394, 2013.

\bibitem{zheng2014bayes}
L.~Zheng, S.~Wang, P.~Guo, H.~Liang, and Q.~Tian,
\newblock ``Bayes merging of multiple vocabularies for scalable image
  retrieval,''
\newblock in {\em IEEE Conference on Computer Vision and Pattern Recognition
  (CVPR)}. IEEE, 2014.

\bibitem{graph}
Z.~Liu, S.~Wang, L.~Zheng, and Q.~Tian,
\newblock ``Visual reranking with improved image graph,''
\newblock in {\em ICASSP}, 2014, pp. 6909--6913.

\bibitem{zheng2014coupled}
L.~Zheng, S.~Wang, Z.~Liu, and Q.~Tian,
\newblock ``Packing and padding: Coupled multi-index for accurate image
  retrieval,''
\newblock in {\em IEEE Conference on Computer Vision and Pattern Recognition
  (CVPR)}. IEEE, 2014.

\bibitem{zheng2014seeing}
L.~Zheng, S.~Wang, F.~He, and Q.~Tian,
\newblock ``Seeing the big picture: Deep embedding with contextual evidences,''
\newblock {\em arXiv preprint arXiv:1406.0132}, 2014.

\bibitem{sivic2003video}
J.~Sivic and A.~Zisserman,
\newblock ``Video google: A text retrieval approach to object matching in
  videos,''
\newblock in {\em IEEE International Conference on Computer Vision}. IEEE,
  2003, pp. 1470--1477.

\bibitem{lowe1999object}
D.~G. Lowe,
\newblock ``Object recognition from local scale-invariant features,''
\newblock in {\em IEEE International Conference on Computer vision}. IEEE,
  1999, vol.~2, pp. 1150--1157.

\bibitem{gray1984vector}
R.~Gray,
\newblock ``Vector quantization,''
\newblock {\em IEEE ASSP Magazine}, vol. 1, no. 2, pp. 4--29, 1984.

\bibitem{shi2012sift}
J.~Shi, Z.~Jiang, H.~Feng, and L.~Zhang,
\newblock ``Sift-based elastic sparse coding for image retrieval,''
\newblock in {\em IEEE International Conference on Image Processing (ICIP)}.
  IEEE, 2012, pp. 2437--2440.

\bibitem{jegou2010aggregating}
H.~J{\'e}gou, M.~Douze, C.~Schmid, and P.~P{\'e}rez,
\newblock ``Aggregating local descriptors into a compact image
  representation,''
\newblock in {\em IEEE Conference on Computer Vision and Pattern Recognition
  (CVPR)}. IEEE, 2010, pp. 3304--3311.

\bibitem{thiagarajan2012supervised}
J.~J. Thiagarajan, R.~K. Natesan, P.~Sattigeri, and A.~Spanias,
\newblock ``Supervised local sparse coding of sub-image features for image
  retrieval,''
\newblock in {\em IEEE International Conference on Image Processing (ICIP)}.
  IEEE, 2012, pp. 3117--3120.

\bibitem{perronnin2010large}
F.~Perronnin, Y.~Liu, J.~S{\'a}nchez, and H.~Poirier,
\newblock ``Large-scale image retrieval with compressed fisher vectors,''
\newblock in {\em IEEE Conference on Computer Vision and Pattern Recognition
  (CVPR)}. IEEE, 2010, pp. 3384--3391.

\bibitem{zhengtopic}
Y.~Zheng, Y.~Zhang, and H.~Larochelle,
\newblock ``Topic modeling of multimodal data: an autoregressive approach,''
\newblock in {\em IEEE Conference on Computer Vision and Pattern Recognition
  (CVPR)}. IEEE, 2014.

\bibitem{liu2012discriminant}
B.~Liu, Y.~Wang, Y.~Zhang, and Y.~Zheng,
\newblock ``Discriminant sparse coding for image classification,''
\newblock in {\em ICASSP}. IEEE, 2012, pp. 2193--2196.

\bibitem{liu2013learning}
B~Liu, Y~Wang, Y~Zhang, and B.~Shen,
\newblock ``Learning dictionary on manifolds for image classification,''
\newblock {\em Pattern Recognition}, vol. 46, no. 7, pp. 1879--1890, 2013.

\bibitem{bo2013multipath}
L.~Bo, X.~Ren, and D.~Fox,
\newblock ``Multipath sparse coding using hierarchical matching pursuit,''
\newblock in {\em IEEE Conference on Computer Vision and Pattern Recognition
  (CVPR)}. IEEE, 2013, pp. 660--667.

\bibitem{jegou2008hamming}
H.~Jegou, M.~Douze, and C.~Schmid,
\newblock ``Hamming embedding and weak geometric consistency for large scale
  image search,''
\newblock in {\em Computer Vision--ECCV 2008}, pp. 304--317. Springer, 2008.

\bibitem{wang2010locality}
J.~Wang, J.~Yang, K.~Yu, F.~Lv, T.~Huang, and Y.~Gong,
\newblock ``Locality-constrained linear coding for image classification,''
\newblock in {\em IEEE Conference on Computer Vision and Pattern Recognition
  (CVPR)}. IEEE, 2010, pp. 3360--3367.

\bibitem{arandjelovic2012three}
R.~Arandjelovic and A.~Zisserman,
\newblock ``Three things everyone should know to improve object retrieval,''
\newblock in {\em IEEE Conference on Computer Vision and Pattern Recognition
  (CVPR)}. IEEE, 2012, pp. 2911--2918.

\bibitem{zheng2013lp}
L.~Zheng, S.~Wang, Z.~Liu, and Q.~Tian,
\newblock ``Lp-norm idf for large scale image search,''
\newblock in {\em IEEE Conference on Computer Vision and Pattern Recognition
  (CVPR)}. IEEE, 2013, pp. 1626--1633.

\end{thebibliography}
}

\end{document}